\pdfoutput=1

\documentclass[11pt]{article}

\usepackage{acl}

\usepackage{times}
\usepackage{latexsym}

\usepackage[T1]{fontenc}

\usepackage[utf8]{inputenc}

\usepackage{microtype}

\usepackage{xargs}                      
\usepackage[colorinlistoftodos,prependcaption,textsize=tiny]{todonotes}

\newcommandx{\unsure}[2][1=]{\todo[linecolor=red,backgroundcolor=red!25,bordercolor=red,#1]{#2}}
\newcommandx{\change}[2][1=]{\todo[linecolor=blue,backgroundcolor=blue!25,bordercolor=blue,#1]{#2}}
\newcommandx{\info}[2][1=]{\todo[linecolor=OliveGreen,backgroundcolor=OliveGreen!25,bordercolor=OliveGreen,#1]{#2}}
\newcommandx{\improvement}[2][1=]{\todo[linecolor=Plum,backgroundcolor=Plum!25,bordercolor=Plum,#1]{#2}}
\newcommandx{\thiswillnotshow}[2][1=]{\todo[disable,#1]{#2}}
\usepackage{booktabs}
\usepackage{multirow}
\usepackage{hyperref}
\newcommand{\userspan}[1]{\emph{``#1''}}
\newcommand{\user}[1]{[\textsc{user:} \emph{``#1''}]}

\newcommand{\slot}[1]{[\texttt{\footnotesize#1}]}
\newcommand{\dst}[2]{[\texttt{\footnotesize #1 = }\emph{``#2''}]}

\newcommand{\cdist}[1]{\delta_{c}}
\newcommand{\triplet}[1]{$\langle $#1$ \rangle$}
\newcommand{\dataset}[1]{\textsc{#1}}
\newcommand{\mwoz}{\dataset{MultiWOZ}}
\newcommand{\sgd}{\dataset{SGD}}
\newcommand{\smcalflow}{\dataset{SMCalFlow}}
\newcommand{\pp}{$pp$}

\usepackage{listings}
\usepackage{algpseudocode}
\usepackage{etoolbox,siunitx}
\robustify\bfseries
\newcommand{\Hquad}{\hspace{0.5em}}

\usepackage[inline]{enumitem}

\title{What Did You Say? 

Task-Oriented Dialog Datasets Are Not Conversational!?}

\author{
    Alice Shoshana Jakobovits \\
    Google \\
    \texttt{jakobovits@google.com} \\\And
    Francesco Piccinno \\
    Google\\
    \texttt{piccinno@google.com} \\\And
    Yasemin Altun \\
    Google\\
    \texttt{altun@google.com} \\
  }

\begin{document}
\maketitle
\begin{abstract}
High-quality datasets for task-oriented dialog are crucial for the development of virtual assistants.
Yet three of the most relevant large-scale dialog datasets suffer from one common flaw: 
the dialog state update can be tracked,
to a great extent, 
by a model that only considers the current user utterance, ignoring the dialog history.
In this work, we outline a taxonomy of conversational and contextual effects, which we use to examine \mwoz{}, \sgd{}~and \smcalflow{}, among the most recent and widely used task-oriented dialog datasets. 
We analyze the datasets in a model-independent fashion
and corroborate these findings experimentally using a strong text-to-text baseline (T5).
We find that less than $4\%$ of \mwoz{}'s turns and $10\%$ of \sgd{}'s turns are conversational,
while \smcalflow{} is not conversational at all in its current release: its dialog state tracking task can be reduced to single-exchange semantic parsing. 
We conclude by outlining
desiderata for truly conversational dialog datasets.
\end{abstract}

\section{Introduction}
\label{sec:introduction}

Virtual assistants such as Alexa, Cortana, Google Assistant and Siri help users carry out all sorts of tasks, ranging from checking the weather and setting alarms to online shopping.
While the development of these agents is a soaring area 
of research, known as task-oriented dialog, their usage lacks in
naturalness, interactivity and in the possibility of accomplishing complex goals requiring rich multi-turn conversation and strong tracking abilities. 

Indeed, in a task-oriented dialog, agents are not only required to carry out tasks specified using single natural language utterances, but they should also be able to interactively combine specifications given over multiple turns (\emph{conversationality}), handling various dialog phenomena such as long and short range references, revisions and error-recovery, while robust to linguistic variations. Beyond the present dialog's history, agents might need to take into account information from an earlier conversation, or even external information that is not explicitly mentioned in the conversation, but should be derived from context (\emph{contextuality}) with varying degrees of ambiguity.


At the crux of these challenges lies the Dialog State Tracking (DST) task. It consists of estimating the dialog state (also known as “belief state”) at a given turn of the dialog.
The DST task raises important questions related to the dialog state's representation 
and the modeling of a conversation's history and context.
To research and answer these questions, however, we need richly \emph{conversational} and \emph{contextual} task-oriented datasets that exhibit these challenges. 
New dialog datasets 
covering more domains, containing longer dialogs and utterances as well as with richer dialog state representations
have been released recently.
But do
these metrics translate to more \emph{conversational} and \emph{contextual} dialogs?

We focus on the two most-cited task-oriented dialog datasets to date (\mwoz{} and \sgd{}), as well as on \smcalflow{} for 
its novel approach of representing dialog state as a graph. 
While varying in terms of data collection methodology, dataset scale and dialog state representation,
our findings show that all three datasets 
lack in conversationality and contextuality.
The contribution of this paper are therefore:

\begin{itemize}[noitemsep,topsep=0pt]
    \item We outline a taxonomy of contextuality / conversationality for dialog datasets (Section~\ref{sec:taxonomy_of_contexts}).
    
    \item We analyze three of the most recent large, multi-domain, task-oriented dialog datasets, (\mwoz{}, \sgd{} and \smcalflow{}) (Section \ref{sec:related-work-and-datasets}) in light of this taxonomy 
    in a model-independent fashion
    (Section~\ref{sec:model_independent_analysis}).
    
    \item We corroborate the model-independent analysis' findings experimentally using T5~\cite{Raffel2019-di} (Section \ref{sec:t5_experiments}). 
    
    \item We show that 
    under $4\%$ of \mwoz's turns are conversational.
    \sgd{} is more conversational (ca. $10\%$), but this is due to annotation policy rather than dialog richness. 
    \smcalflow{} is non-conversational in its current setup and dataset release, and is akin to a single-exchange non-conversational semantic parsing dataset than to a dialog dataset.
    

\end{itemize}

\section{Related Work \& Datasets}
\label{sec:related-work-and-datasets}


\paragraph*{\mwoz{}}
%

Collected through a Wizard-of-Oz
process~\cite{Kelley1984-hx}, \mwoz~\cite{Budzianowski2018-oa} was a breakthrough dataset for task-oriented dialog research. At about an order of magnitude larger than the task-oriented dialog datasets available thus far and featuring $7$ task domains as well as over $7$\si{\kilo} multi-domain dialogs (Table~\ref{tbl:datasets}), \mwoz{} became a standard benchmark for various dialog tasks including DST.
\paragraph*{\sgd{}} (Schema Guided Dataset)~\cite{Rastogi2019-kd}
features a much larger number of domains than \mwoz{} and several different services (or schemas) for a given domain by using a dialog simulator to generate templates of dialogs including dialog state and asking crowdworkers to formulate these structures into natural language. It is the most cited task-oriented dialog dataset after \mwoz{} to date.


\paragraph*{\smcalflow{}}\cite{Andreas2020-xl}
provides a richer representation of the dialog state than 
the semantic frames (a structured intent-slot-value list) employed in \mwoz{} and \sgd{}:
it is a dataflow graph, equivalently expressed as a program in Lispress (a programming language proposed by the datasets' authors) which fulfills the user's request. The graph representation can potentially create opportunities to capture richer and more complex dependencies throughout the dialog, and in turn provide more extensive context modeling explorations. 
Its dialog states also feature explicit functions for references and revisions.



\begin{table}[ht]
\footnotesize
\begin{tabular}{@{}l|p{1.9cm}p{2.4cm}p{2cm}@{}}
\toprule
\textbf{}        & \textbf{\mwoz{}}    & \textbf{\sgd{}}            & \textbf{\smcalflow{}} \\ \midrule
\parbox[t]{1mm}{\multirow{3}{*}{\rotatebox[origin=c]{90}{\textbf{Size}}}}

                 & $10$\si{\kilo} dialogs   & $22$\si{\kilo} dialogs  & $40$\si{\kilo} dialogs \\
                 & $155$\si{\kilo} turns    & $460$\si{\kilo} turns   & $312$\si{\kilo} turns \\
                 & $7$ domains              & $18$ domains,           & $4$ domains \\
                 &                          & $26$ services    &  \\
\midrule
\parbox[t]{1mm}{\multirow{3}{*}{\rotatebox[origin=c]{90}{\textbf{Collect.}}}}
    & WoZ \newline CS annotation
    & Simulator \newline Templated text gen. \newline CS paraphrasing
    & WoZ~\newline CS annotation \\
\midrule
\parbox[t]{1mm}{\multirow{2}{*}{\rotatebox[origin=c]{90}{\textbf{DST}}}}
                    & Semantic frame & Semantic frame & DS as a graph \\
                    & User-centric      & User-centric      & Shared \\
\bottomrule
\end{tabular}
\caption{\label{tbl:datasets}Datasets statistics.
DS stands for dialog state, WoZ for Wizard-of-Oz, CS for crowd-sourced.
}
\end{table}

\paragraph*{Context Modeling for Dialog}

The research into dialog history modeling is much more extensive for open-domain dialog 
than task-oriented dialog~\cite{tian-etal-2017-make}. 
In the former, 
dialog history representation has been explored by, e.g., representing the entire dialog history as a linear sequence of tokens~\cite{sordoni-etal-2015-neural}, using a fixed-size window to represent only the recent dialog history~\cite{li-etal-2016-diversity}, designing hierarchical representations~\cite{serban2016building,xing2017hierarchical,shen-etal-2019-modeling,zhang-etal-2019-recosa}, leveraging structured attention~\cite{qiu-etal-2020-structured,su-etal-2019-improving} as well as summarizing~\cite{xu2021goldfish} or re-writing~\cite{xu-etal-2020-semantic} dialog history to handle long dialogs.

The literature around dialog history modeling for task-oriented dialog on the other hand is sparser: many DST models are supplied only with the last exchange utterances, i.e., the last agent and the current user utterance \cite{Rastogi2019-kd, Andreas2020-xl}, some even with just the current user utterance~\cite{platanios-etal-2021-value}. 
Cross-domain transfer~\cite{wu-etal-2019-transferable}, slot-correlations~\cite{Ye2021-af}, pre-training~\cite{zhao-etal-2021-effective-sequence} are topics that have been explored for DST modeling. 
In this paper, we focus however on the datasets and investigate their conversationality/contextuality. 



\section{A Taxonomy of Dialog: Conversationality \& Contextuality}
\label{sec:taxonomy_of_contexts}

We define a dialog as a succession of written natural language utterances 
in which a user (i.e., a human with one or multiple intended goals) and an agent (i.e., an automated system tasked with fulfilling these goals) take turns contributing to the dialog. Each participation is therefore called a \emph{turn}, and we define a pair made up of an agent turn and its consecutive user turn as one \emph{exchange}.

\begin{figure}[t]
    \centering
    \includegraphics[width=0.90\linewidth]{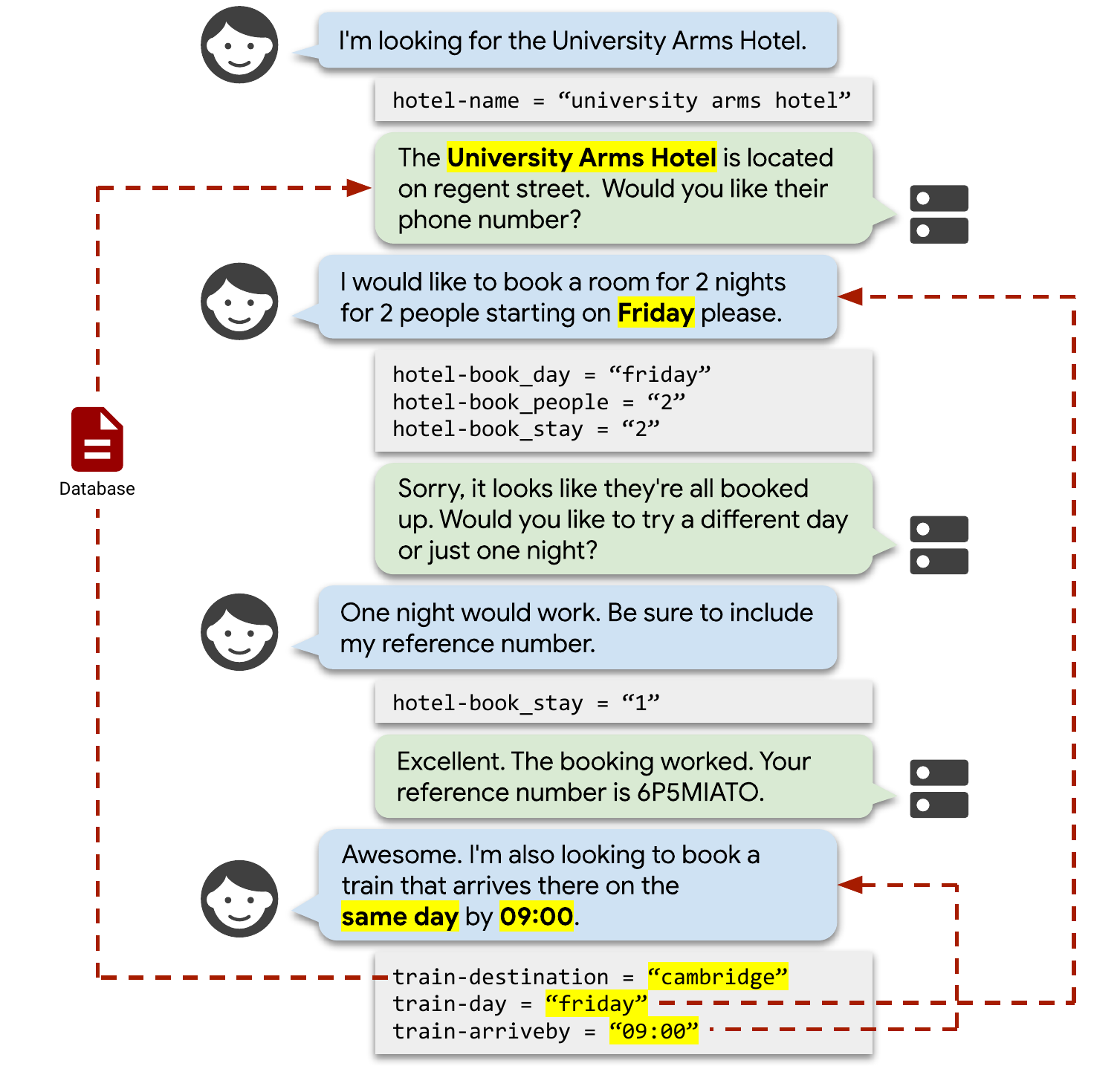}
    \caption{Example~\texttt{MUL0635} from \mwoz{}. 
    The bottom box shows the dialog state update at the current turn, the arrows indicate the origin of each slot value. Here, \userspan{cambridge} has $\cdist{}=5$ and requires world knowledge, \userspan{friday} has $\cdist{}=4$ and \userspan{09:00} of $\cdist{}=0$.
    \sgd{} exhibits a similar dialog state structure.
    }
    \label{fig:taxonomy_of_contexts:example:multiwoz}
\end{figure}

\paragraph*{Conversationality}
We define a given turn in a dialog as \textit{conversational in the DST task} if the dialog history (i.e., the turns prior to the current exchange) is required for correctly tracking the current turn’s dialog state. 
In other words, in a truly conversational dialog, utterances cannot be parsed in isolation, because their meaning is highly dependent on what was said in previous turns. 
We quantify this property using the conversational distance $\cdist{}$, which measures the number of turns a system has to look back into the dialog history to accurately predict a given turn's dialog state update.
Slots whose values are found in the current user turn (i.e., the turn immediately preceding the dialog state, such as the \slot{train-arriveby} slot in Figure~\ref{fig:taxonomy_of_contexts:example:multiwoz}) are defined as \emph{non-conversational}, as they have a conversational distance $\cdist{} = 0$.
We also consider slots whose values are found in the last agent turn (i.e., at a conversational distance of $\cdist{} = 1$) as \emph{non-conversational}. 
Indeed, they too belong to the current exchange; as such, their value is only represented in the previous agent utterance, but not yet recorded in the dialog state at this stage of the conversation.
By contrast, slots such as \slot{train-day} in Figure~\ref{fig:taxonomy_of_contexts:example:multiwoz} with $\cdist{} = 4$ are regarded as \emph{conversational}.
Finally, a \emph{turn} has a $\cdist{}$ equal to the maximum $\cdist{}$ of all slot-values in its dialog state update.
A robust DST system should support such conversational effects for both short and long ranges, as well as revisions and ambiguities in references. In the extreme case, DST could even depend on information provided in past conversations, as is the case for \textit{Multi-Session Chat} open-domain dialog dataset~\cite{xu2021goldfish}. 

\begin{figure}
    \centering
    \includegraphics[width=0.84\linewidth]{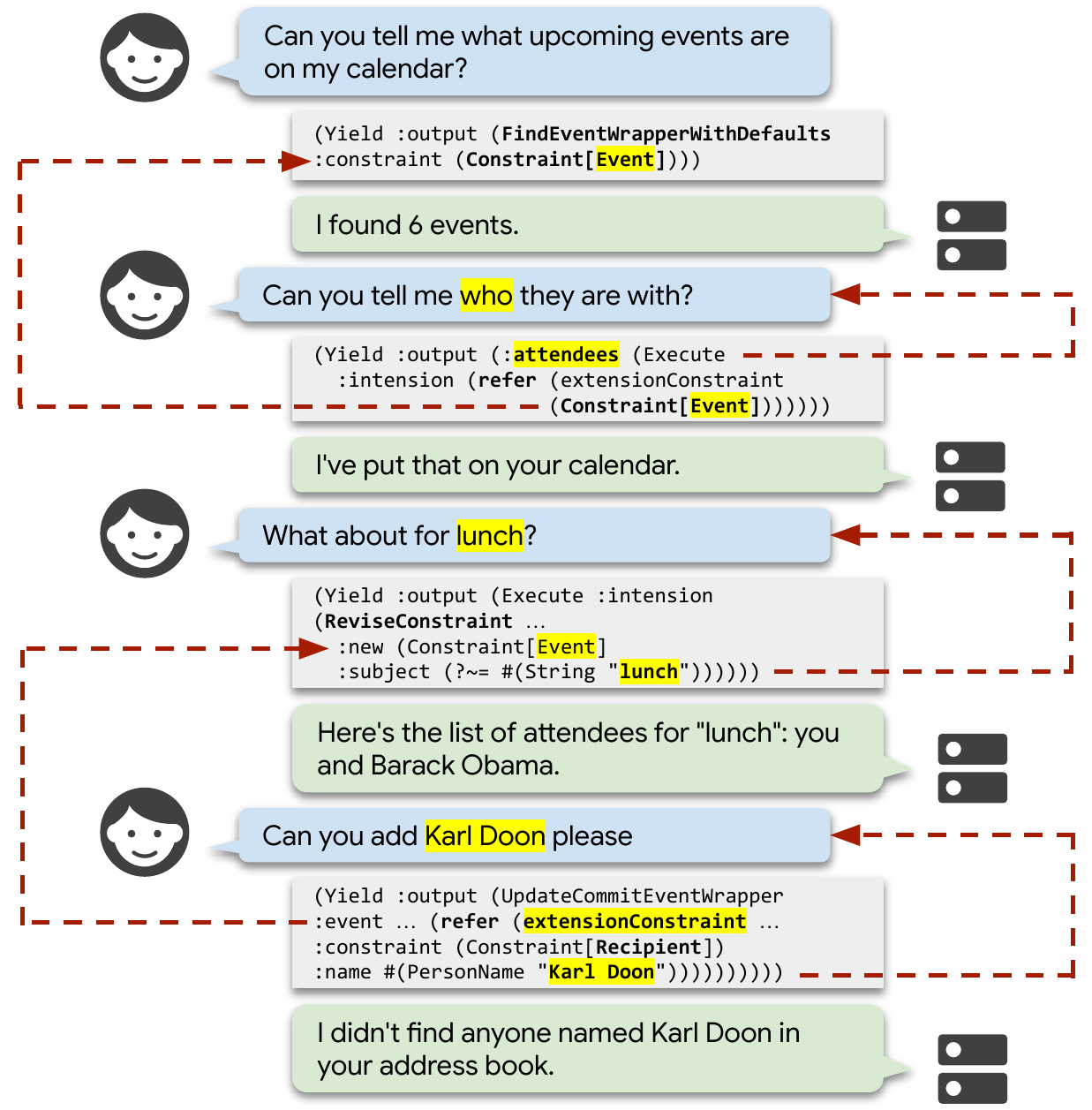}
    \caption{Dialog example from the \smcalflow{} dataset (\texttt{\footnotesize dee6e250-\-bcc9-\-47e7-\-98fc-\-6b589c234868}). 
    The dialog state is given in Lispress graph program form at each turn of the dialog.}
    \label{fig:taxonomy_of_contexts:example:smcalflow}
\end{figure}

\paragraph*{Contextuality} 
We define a given turn in a dialog as \textit{contextual in the DST task} if the dialog state at a given turn is dependent not only on the dialog history, but also on elements beyond the conversation itself, that are not explicitly mentioned in the conversation. These elements could be: 

\begin{description}[style=unboxed,leftmargin=0cm]

\item[\raisebox{.5pt}{\textcircled{\raisebox{-.9pt} {1}}} Situational:] the slot value in the dialog state depends on the circumstances of the dialog, e.g., its date or location, that are not explicitly mentioned. 
For example, in the utterance: \user{Hey, I feel like listening to some tunes right now. Can you find me something from two years ago?}, the dialog state tracker must recognize that the current year is $2022$ in order to track the state \dst{year}{2020}.

\item[\raisebox{.5pt}{\textcircled{\raisebox{-.9pt} {2}}} Knowledge about the user:] the dialog state depends on some knowledge about the user (e.g., dietary restrictions or movie preferences).

\item[\raisebox{.5pt}{\textcircled{\raisebox{-.9pt} {3}}} External knowledge:] the slot value in the dialog state depends on some world knowledge and requires, e.g., a query from an external database. In the dialog in Figure~\ref{fig:taxonomy_of_contexts:example:multiwoz}, for example, the dialog state tracker must deduce \dst{train\_destination}{cambridge} by understanding that \userspan{there} refers to \userspan{University Arms Hotel}, then querying some external database to get the train stop corresponding to this location. Note that the value \userspan{cambridge} is never mentioned explicitly in the dialog.

\end{description}

\paragraph*{Dialog state value normalization}
In addition to conversationality and contextuality, a robust DST system must be robust to linguistic variations, meaning it should recognize semantically equivalent expressions and convert them into a normalized form. 
Examples of such normalization include:
\begin{enumerate*}[label=(\roman*)]
    \item \textit{Typos} (e.g., \userspan{18:!5} $\rightarrow$ \dst{time}{18:15}),
    \item \textit{Entity Recognition} (e.g., 
    \userspan{thirty bucks} $\rightarrow$ \dst{price}{\$30})
    \item \textit{Semantic Understanding} (e.g., \userspan{on a budget} $\rightarrow$ \dst{price\_range}{inexpensive}), and
    \item \textit{Computation} 
    (\userspan{from tuesday through thursday} $\rightarrow$ \dst{book\_stay}{2})\footnote{Refer to Table~\ref{tbl:slot_value_normalization} in the Appendix for a more detailled list and additional examples.}.
\end{enumerate*}
These effects can also occur in combination with one another.

Some of these linguistic variations could be handled using components developed independently, e.g.~auto-corrects or entity linkers, and incorporated into dialog systems. Still, the challenge of how to recognize these values as semantically equivalent and how to normalize them accurately must be addressed by robust agents and must therefore be represented in research datasets.

\section{Model-independent Analysis}
\label{sec:model_independent_analysis}

\begin{table}[ht]
\centering
\footnotesize
\begin{tabular}{@{}lS[table-format=2.3]S[table-format=2.3]|S[table-format=3.2]cc@{}}
\toprule
 & \multicolumn{2}{c}{\textbf{\mwoz{}}}      & \multicolumn{1}{c}{\textbf{\sgd{}}} \\
\midrule
\textbf{Conversationality} 
 & \textsc{test} & \textsc{dev} & \multicolumn{1}{c}{\textsc{test}} \\
\Hquad nothing to predict
 & 32.63         & 31.15        & 45.66 \\
\quad $+ \cdist{} = 0$
 & 85.75         & 85.83        & 80.91     \\
\quad $+ \cdist{} = 1$
 & 96.48         & 96.26        & 90.42 \\
\midrule
\Hquad $\cdist{} \geq 2$
 & 3.52          & 3.64         & 9.58 \\
\midrule
\textbf{Contextuality} 
 & \textsc{test} & \textsc{dev} & \multicolumn{1}{c}{\textsc{test}} \\
\Hquad non-contextual
 & 99.96    & 99.96   & 100 \\
\Hquad situational
 & 0.01     & 0.03    & 0 \\
\Hquad knowledge about the user
 & 0        & 0       & 0 \\
\Hquad external knowledge
 & 0.03     & 0.01    & 0 \\
\midrule
\textbf{Normalization} 
 & \textsc{test} & \textsc{dev} & \multicolumn{1}{c}{\textsc{test}} \\
\Hquad verbatim
 & 87.30    & 87.40   & 93.92 \\
\Hquad typos
 & 2.14     & 2.52    & 0 \\
\Hquad semantic understanding
 & 5.12     & 4.95   & 6.49 \\
\Hquad computation
 & 0.08     & 0.05    & 0.11 \\
\Hquad other 
 & 5.86    & 5.64   & 3.72 \\
\bottomrule
\end{tabular}
\caption{\label{tbl:model_independent_analysis}Model independent analysis: \% of turns in the \mwoz{} and \sgd{} datasets that feature effects from the taxonomy
defined in Section~\ref{sec:taxonomy_of_contexts}. The percentages of normalization effects add up to $>$100\% since there are turns that feature multiple different normalization effects.
}
\end{table}



\paragraph{\mwoz{} \& \sgd{}} \noindent
With the taxonomy of conversational and contextual effects at hand (Section~\ref{sec:taxonomy_of_contexts}), we analyze each turn of the \mwoz{} version $2.4$~\cite{Ye2021-jo} (dev. and test sets\footnote{Our analysis on a sample of training data shows similar characteristics in training sets.}) and \sgd{} (test set\footnotemark[\value{footnote}]) dialogs semi-automatically\footnote{The pseudocode can be found in Figure~\ref{fig:model_independent_analysis:pseudocode} of the Appendix. We plan to open-source the associated scripts.}.
The results
are summarized in Table~\ref{tbl:model_independent_analysis}. 


In terms of \textbf{conversationality}, 
over $85\%$ of \mwoz{}'s turns and over $80\%$ of \sgd{}'s turns
have either 
(i) an empty dialog state update (i.e., nothing to predict) 
or 
(ii) a dialog state update that can be predicted by considering only the current user turn (i.e., $\cdist{}=0$). 
If we further include the last agent turn (i.e., $\cdist{}=1$), over $96\%$ of \mwoz{}'s turns and over $90\%$ of \sgd{}'s turns can be predicted correctly without using the latest dialog state and/or its corresponding information,
and are therefore \emph{non-conversational}. 
This leaves only under $4\%$ (\mwoz{}) and $10\%$ (\sgd{}) of turns with a conversational distance above $\cdist{} \geq 2$, which require looking at a conversational window beyond the most recent exchange for their dialog state to be tracked correctly. 
\mwoz{} is therefore much less conversational than \sgd{}.

However even for \sgd{}, information beyond the latest exchange is irrelevant to the dialog state update most of the time.
This lack of conversationality can be attributed, at least in part, to the dataset's design: in the data collection procedure, crowd-workers are asked to paraphrase dialog structures, as generated by a dialog simulator, into natural language by writing out the current slot values verbatim, and without resorting, for example, to shorthand references (such as \userspan{that}, \userspan{the first one}). One might have thought that this strategy would entirely eliminate the reference issue, and render the dataset non-conversational, but that is not the case. To understand why, let us look at the distribution of conversational distances in \mwoz and \sgd{}'s conversational slices.

\begin{figure}[ht]
    \centering
    \includegraphics[trim=35 10 0 24, clip,width=0.9\linewidth]{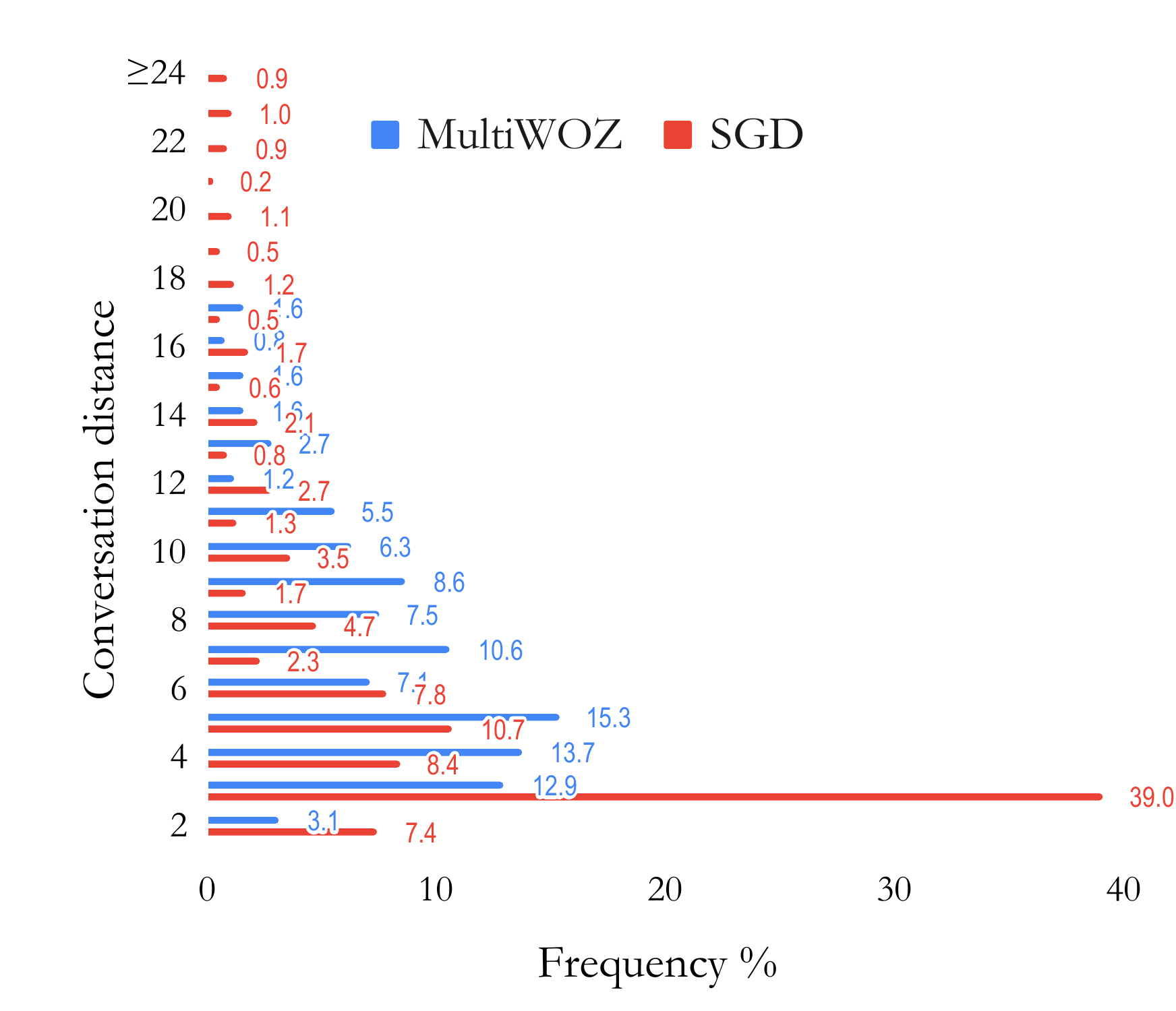}
    \caption{Distribution of the conversational distances $\cdist{}$ of turns from \mwoz{} and \sgd{}'s test set. For scale, we represent only the turns with $\cdist{} \geq 2$.}
    \label{fig:model_independent_analysis:reference_range_histogram}
\end{figure}

Figure~\ref{fig:model_independent_analysis:reference_range_histogram} shows the distribution of turns' conversational distances in the conversational slices for the two datasets. The maximum $\cdist{}$ in \mwoz{} is of $17$, while \sgd{} reaches further, with $\cdist{}$ of more than $24$.
The majority of \mwoz{}'s references are approximately evenly distributed at distances between $3$ and $5$, whereas \sgd{} features a peak at $\cdist{}=3$. 
This is due to a frequent dialog structure in \sgd{} that explains its conversationality and can be summarized as follows: (i) at $\cdist{}=3$, the agent asks for a confirmation before booking (this utterance contains information relevant to the dialog state, but the dialog state is not updated); (ii) at $\cdist{}=2$, the user asks a clarification question; (iii) at $\cdist{}=1$ the agent answers, (iv) then the dialog state is updated at $\cdist{}=0$, upon user confirmation, with values that were stated by the agent at $\cdist{}=3$. 
This patterns follows from a dataset design decision, where the dialog state is updated only following an \texttt{INFORM}, \texttt{SELECT}, \texttt{AFFIRM} or \texttt{NEGATE} intent by the user.
In contrast, in \mwoz{}, if a slot value is mentioned by the agent, and is not confirmed nor rejected by user, the dialog state is nevertheless updated immediately\footnote{See dialog \texttt{PMUL3897}, turn~$8$, in \mwoz{}, for example, where the user asks for more information before booking.}.
Hence the conversationality of a dataset results not only from the richness of its dialogs and natural language utterances, but to a large extent from the annotation policy chosen. 

Lastly, it is noteworthy that while the value of a slot can be changed from one exchange to the next (e.g. \dst{area}{center} at turn $T-2$ and \dst{area}{north} at turn $T$), it is only very rarely dropped or changed to \emph{``dontcare''}. In fact, this happens in $2.08\%$ of turns for \mwoz{} and $0.27\%$ of turns for \sgd{}.
In other words, constraints expressed in the dialog state are sometimes changed, but a constraint is almost never relaxed, which is why a DST system trained on these datasets cannot learn when to remove slot-values from the dialog state, only when to add.

In terms of \textbf{contextuality}, Table~\ref{tbl:model_independent_analysis} showcases that the overwhelming majority of both datasets turns ($99.96\%$ for \mwoz{}, $100\%$ for \sgd{}) are \emph{non-contextual}, in that the dialog state of these datasets can be estimated by and large only by looking at the dialogs themselves, without taking circumstances, knowledge about the user or world knowledge into account. In that respect, they fall short of conversing in the way humans do: assuming knowledge about the world, themselves, and the circumstances of the conversation.


In terms of \textbf{value normalization} (Table~\ref{tbl:model_independent_analysis}), an overwhelming majority of slot values can be found verbatim in the dataset ($87.30\%$ for \mwoz{}, $93.92\%$ for \sgd{}), limiting the possibility of evaluating DST systems' robustness to linguistic variations. The \emph{semantic understanding} slice, which is at the core of semantic parsing (be it conversational or non-conversational), is of $5.12\%$ and $6.49\%$ respectively. Arguably the most challenging of these effects, namely turns requiring computation, are negligible (ca. $0.1\%$). 
For \mwoz{}, the proportion of non-verbatim turns ($12.70\%$) is much larger than the proportion of conversational turns ($3.52\%$), 
indicating that value normalization effects are more predominant than conversational and contextual effects in this benchmark.
This suggests that \mwoz{} measures a model's robustness to lexical variation more than it measure a model's conversational capacities.


\paragraph{\smcalflow{}} \noindent
Contrary to \mwoz{} and \sgd{}, the dialog state of \smcalflow{}
does not take the form of a semantic frame (a structured intent-slot-value list).
Instead, an exchange between user and agent is conceptualized as follows:
(i) the user formulates an utterance in natural language, 
(ii) the system predicts a Lispress program 
which, when executed, will fulfill the user's request 
and is formally a dataflow graph representing a shared belief of the state of the dialog, 
(iii) the program is evaluated, and the results are added as nodes extending the dataflow graph, 
and (iv) the agent's natural-language response is generated.

The graph/program formalism provides opportunities to capture complex tasks, including modeling of compositionality across domains, intents and/or arguments as well as conversational phenomena such as reference and revisions with non-trivial dependencies. Indeed, the dialogs are quite conversational:
the percentage of turns featuring a reference is $29.19\%$ (Lispress program with call to function \texttt{refer()}) and the percentage of turns featuring a revision (Lispress program with \texttt{revise()}) is $8.77\%$. However, the API of these functions do not contain the resolved referents but only an optional constraint on the type, property or role of the referred-to object. The DST task, as conceptualized in \smcalflow{}, does not entail reference or revision resolution. Rather these must be obtained through a separately-trained saliency model. This formalism is unobjectionable in and of itself, but because the evaluated programs (step (iii) described above) are missing from the public release, the data that would be needed to train such a saliency model is not available to the research community. In a nutshell, the DST task simply requires to predict the correct \texttt{refer()} or \texttt{revise()} call, without resolving it, and the task therefore renders to non-conversational semantic parsing.  

Similarly, while \smcalflow{} has the potential to provide rich contextuality, particularly in user context with calendar entries and contacts, the unavailability of this knowledge base in its public release renders the data non-contextual for researchers at large.

\section{T5 Experiments}
\label{sec:t5_experiments}

\begin{table}[ht]
\footnotesize
\centering
\begin{tabular}{@{}l|cccc@{}}
\toprule
\textbf{Input representation} & \multicolumn{3}{c}{\textbf{Linearization}} \\
\midrule
\Hquad current user turn
&
& 
& 
& $U_T$ \\
\quad $+$ last agent turn
&
&
& $A_{T-1}$ 
& $U_T$ \\
\quad $+$ previous dialog state
&
& $S_{T-2}$ 
& $A_{T-1}$ 
& $U_T$ \\
\Hquad full dialog history 
& $U_0$ $A_1$
& $\dots$ 
& $A_{T-1}$ 
& $U_T$ \\
\bottomrule
\end{tabular}
\caption{Linearizations of the dialog for a given user turn $T$ and different input representations exhibiting more or less conversational context. In the table, $U_T$, $A_T$ and $S_T$ stand respectively for user utterance, agent utterance and linearized state at turn $T$.}
\label{tbl:t5_linearizations}
\end{table}

In this section, we run experiments with a strong text-to-text baseline, T5~\cite{Raffel2019-di}, to investigate how and to what extent the findings from the model-independent analysis (Section~\ref{sec:model_independent_analysis}) are reflected in T5 models. 

\subsection{Method}
\label{sec:t5_experiments:method}

Traditionally, in DST for task-oriented dialog, intent prediction has been framed as a classification task and slot-value prediction as a span labelling task~\cite{Rastogi2019-kd,Chen2020-wx}. However, 
recent works have explored DST in the seq2seq setting~\cite{wen-etal-2018-sequence, Gao2019-qt, Feng2021-tb}. 
We follow this formalization 
as it is flexible enough to generalize to multi-intent and multi-domain utterances, as well as to dialog state representations more complex than intent-slot-value triplets, such as Lispress programs.

\paragraph*{Model}

We use the publicly released T5 1.1 base checkpoint\footnote{\url{https://github.com/google-research/t5x}, $\sim$250 million parameters}~
and fine-tune it
on
each of 
the three datasets studied here.~
We use a learning rate of $1\text{e-}3$ and a batch size of $128$. We adapt input and output sequence length 
so that the entirety of the longest data sample 
can be represented\footnote{Refer to Table~\ref{tbl:t5_hyperparameters} in the Appendix for additional details on the experimental setup.}, and use default values for all other hyperparameters. 


\paragraph*{Input representations}
To explore how conversational windows of different widths affect the model's performance, we run experiments with different input representations that include distinct levels of conversational context, as shown in Table~\ref{tbl:t5_linearizations}. We also explore the use of the dialog state as a summary of the dialog history thus far.

\paragraph*{Linearization}
For \mwoz{}, the dialog state update is a list of \triplet{domain, slot name, slot value} triplets, which we linearize as a comma-separated list of strings ``\texttt{\footnotesize domain:slot\_name=slot\_value}''. The input is linearized by concatenating the utterances (and states) according to Table~\ref{tbl:t5_linearizations}, and prepending the tag ``\texttt{\footnotesize [user]}'', ``\texttt{\footnotesize [agent]}'' or ``\texttt{\footnotesize [states]}''. 
For \sgd{}, we follow previous work by prepending service schema descriptions to the input 
. 
In \smcalflow{}, the dialog state update is a Lispress program, and is therefore already linearized. Our input linearization is the same as for \mwoz{} and \smcalflow{}, except we use the same tags as the dataset authors (``\texttt{\footnotesize \_\_User}'', ``\texttt{\footnotesize \_\_Agent}'' and ``\texttt{\footnotesize \_\_State}'') for a fair comparison.  

\paragraph*{Evaluation metrics}
For \mwoz{} and \sgd{}, 
we report Joint Goal Accuracy (JGA), 
using the TRADE~\cite{wu-etal-2019-transferable} evaluation script\footnote{\url{https://github.com/jasonwu0731/trade-dst}} and the DSTC8 evaluation script\footnote{\url{https://github.com/google-research/google-research/tree/master/schema_guided_dst}}, respectively. We report both oracle (i.e., the predicted state update is added to the \emph{gold} previous state) and non-oracle (i.e., the predicted dialog state update at turn $T$ is added to the \emph{predicted} state at turn $T-1$) results. 
For \smcalflow{}, 
we use the evaluation script published by the datasets' authors\footnote{\url{https://github.com/microsoft/task_oriented_dialogue_as_dataflow_synthesis}} and report exact-match accuracy. \smcalflow{}'s test split has not been made public, which is why we train on the training split and evaluate on the validation split.

\subsection{Results}
\label{sec:t5_experiments:results}

The results obtained from the T5 model are shown in Table~\ref{tbl:t5_experiments}. 
Non-oracle JGA results are given for \mwoz{} and \sgd{} to allow for comparison (merely for the sake of reference) with previous state-of-the-art results on these datasets.
While non-oracle JGA is more true to the actual performance of a DST system, it has the disadvantage of introducing numerous accumulation errors which make it difficult to isolate the reason for a model's failing.
Since our objective is not to evaluate the performance of a given DST system against previous approaches, but rather to evaluate the conversationality of widely-used DST benchmarks, we use \emph{oracle} JGA for the purposes of our work. It allows us to zero in on the turn a system didn't track right, preserving just the original error and eliminating propagation effects.


The results in Table~\ref{tbl:t5_experiments} experimentally confirm the conclusions drawn from the model-independent analysis: namely that \textit{these datasets can be solved, to a large extent, by showing the model only the current user utterance}: we obtain over $63\%$ accuracy in \sgd{} and over $70\%$ in \mwoz{} and \smcalflow{} by feeding the model only with the current user utterance and no further context or dialog history at all. Moreover, the percentage point improvements brought by each wider conversational window are in line with the model-independent analysis: for \mwoz{}, results are consistently around $10\%$ below the proportion found to be solvable at this $\delta_c$ for each conversational window. For \sgd{}, results are consistently around $20\%$ below the proportion found to be solvable at this $\delta_c$. The lower accuracy on \sgd{} is unsurprising since \sgd{} is a harder benchmark than \mwoz{} due to its more extensive ontologies, numerous services and existence of multiple services per domain.



\begin{table}[ht]
\centering
\footnotesize
\begin{tabular}{@{}lS[table-format=2.2]S[table-format=2.2]S[table-format=2.2]@{}}
\toprule
\textbf{Input representation} &
\rotatebox{45}{\textbf{\footnotesize{\mwoz{}}}} &
\rotatebox{45}{\textbf{\footnotesize{\sgd{}}}} &
\rotatebox{45}{\textbf{\footnotesize{\smcalflow{}}}} \\

\midrule
& \multicolumn{2}{c}{\textsc{~~~oracle jga}} &  \textsc{~~~~~ex. m.}\\

\midrule
\Hquad current user turn
 & 77.19    & 63.20 & 71.53 \\
\quad $+$ last agent turn
 & 85.71    & 69.89 & 78.50 \\
\quad $+$ previous d. state
 & 89.13    & 73.06 & 79.13 \\
\Hquad full dialog history
 & 89.92    & 78.07 & 79.10 \\

\midrule
& \multicolumn{2}{c}{\textsc{~~~~~jga}}  &  \\
\midrule

\Hquad current user turn
 & 43.00    & 22.96 &  \\
\quad $+$ last agent turn
 & 59.43    & 28.66 &  \\
\quad $+$ previous d. state
 & 66.16    & 35.25 &  \\
\Hquad full dialog history
 & 68.91 & 43.65 &  \\

\midrule
\Hquad SOTA
 & {\bfseries 73.62}
 & {\bfseries 73.75}  
 & {\bfseries 80.4} \\

\bottomrule
\end{tabular}






\caption{Dialog State Tracking (DST) results in a seq2seq setup with T5 in \% with different input representations (see Table~\ref{tbl:t5_linearizations}). We report Joint Goal Accuracy (JGA, oracle and non-oracle), on \mwoz{}'s and \sgd{}'s test set, and exact-match accuracy (Ex. M.) on \smcalflow{}'s validation set (unseen during training). Each score corresponds to a single run. 
For reference, 
state-of-the-art results on \mwoz{}~\cite{Ye2021-af,Ye2021-jo}, \sgd{}~\cite{Ruan2020-br} and \smcalflow{}~\cite{platanios-etal-2021-value} are shown.
}
\label{tbl:t5_experiments}
\end{table}

\paragraph*{\mwoz{}}

Overall, the T5 model for \mwoz{} does benefit from being trained on dialog history beyond just the last user turn.
However, most of the percentage point improvement comes from adding the last agent utterance ($+8.52$~\pp), while including the previous dialog state or the full dialog history bring much more modest improvements of $+3.42$ or $+4.21$~\pp. This is consistent with the findings from the model independent analysis, which showed that many more turns require slot values to be retrieved from $\delta_c = 1$ ($10.73\%$) than from $\delta_c \geq 2$ ($3.52\%$).

From the $+8.52$~\pp{} improvement brought by adding the last agent utterance, the majority ($74\%$) of turns are indeed turns of $\cdist{}=1$, where one of the slot values, names or domains is stated by the agent. 
The remainder is mostly cases involving the \slot{hotel-type}, \slot{leaveat} or \slot{arriveby} slots: ``\texttt{\footnotesize hotel}'' is both a domain and a candidate value of the \slot{hotel-type} slot, leading to confusions
; \slot{leaveat} or \slot{arriveby} slots should only be updated updated when they refer to a requested time, not a concrete timetable reading, which is sometimes unclear from the current user utterance alone.
The improvements brought from including the previous dialog state or the full dialog history are due to information at $\cdist{} \geq 2$ to a slightly lesser extent: $48\%$ and $68\%$, respectively. The leftover errors are most often a missing \slot{name} slot of an attraction, restaurant or hotel
.

\paragraph*{\sgd{}}
For \sgd{}, providing the model with the last agent utterance similarly improves JGA by $+6.69$~\pp, and providing the previous dialog state in addition improves by $+3.17$~\pp. However, training a model with the full dialog history improves over the last-exchange baseline by a much larger $+8.18$~\pp. This can be explained by the fact that \sgd's conversational slice is a lot larger than \mwoz's ($9.58\%$ vs. $3.52\%$, see Table~\ref{tbl:model_independent_analysis}) and by the fact that in \sgd{}'s conversational slice, most slots are found at $\delta_c = 3$ (see Figure~\ref{fig:model_independent_analysis:reference_range_histogram}), the values of which can only be retrieved by the full dialog history model, not by the previous-state model. 

Similarly to \mwoz{}, not all improvements brought by wider conversational windows are due to conversationality. In fact, $48\%$, $57\%$ and $61\%$ of the improved-upon turns in the last agent, previous dialog state and full history, respectively, are due to the surfacing of information of corresponding $\cdist{}$. The most prominent other source of errors is slot name confusion, whereby a slot value is predicted correctly, but there is a confusion between two schemas, e.g. \dst{city}{danville} vs. \dst{location}{danville}.



\paragraph*{\smcalflow{}}
Similarly to \sgd{}, the ``current user turn'' baseline has a very high accuracy ($71.53\%$). While this baseline is significantly improved upon by showing the model the last agent utterance ($78.50\%$, i.e. $+6.97$\pp), adding the previous dialog state ($79.13\%$, i.e. $+0.63$\pp) or the full dialog history ($79.10\%$, i.e. $+0.60$\pp) bring only marginal improvements. 
The T5 experiments therefore confirm the finding from the model-independent analysis, namely that since reference and revision mechanisms must simply be predicted by an API call at DST time and not actually resolved, the DST task as formalized in this dataset is inherently non-conversational and can be reduced to a semantic parsing task.

This lack of conversationality is implicitly given away in the dataset paper~\cite{Andreas2020-xl}: there, the authors explore contexts of conversational distance $\cdist{}=0$, $1$ and $2$, then use a context window of~$1$ because it gives them the best results. In their follow-up paper~\cite{platanios-etal-2021-value} however, the authors completely ignore dialog history and train using the last user turn exclusively, and in doing so, they obtain a better per-turn exact match accuracy of~$80.4\%$. In both cases, they do not use any contextual information. The fact that they obtain such a high accuracy without any conversational modeling improvements, by only showing the model the latest user turn implies that the dataset is not conversational nor contextual. 

Furthermore, error analysis reveals the presence of a questionable type of error: 
our model predicts the program entirely correctly for certain turns, which are nevertheless evaluated to an accuracy of $0$ by the authors' evaluation script. 
These turns are marked in the dataset as \texttt{refer\_are\_incorrect}, and because of that, they are scored with an accuracy as $0$, no matter the accuracy of the predicted program.
Hence, that the program prediction task cannot be solved entirely on this dataset using the setting proposed by its authors.



\section{Conclusion}
\label{sec:conclusion}


In this work, we outlined a taxonomy of conversational, contextual and linguistic normalization effects that a robust dialog state tracking system should support. 
We evaluated three recent large-scale task-oriented dialog datasets (\mwoz{}, \sgd{}, \smcalflow{}) against this taxonomy in a model-independent fashion and found that both \mwoz{} and \sgd{} exhibit a low rate of conversational turns (under $4\%$ and $10\%$, respectively). 
The majority of \sgd{}'s conversational turns have a conversational distance $\cdist{}=3$.
We showed this is due to \sgd{}'s annotation policy rather than to the inherent richness of its dialogs. 
Though \smcalflow{} prominently features conversational effects such as references ($29.19\%$) and revisions ($8.77\%$), 
its conversational effects are abstracted away from the DST task and the dataset's public release does not feature all the elements needed (i.e. evaluated programs or an execution module for Lispress) for the wider community to investigate
the modeling of said references and revisions. 
\smcalflow{}'s DST task is therefore non-conversational in its current setup and release, and can be reduced to a single-exchange non-conversational semantic parsing task.
Finally, we corroborated these findings experimentally with a strong text-to-text baseline. 

We limited the scope of this work to three datasets, but for completeness, more datasets (in particular non-english, multi-modal and spoken datasets) should be studied. Moreover, we focused on the DST task while a complete study of dialog would have to examine the conversationality and contextuality of the response generation task as well, for instance.

To advance the state-of-the-art in task-oriented dialog research, dataset design and collection procedures may increase focus on
(i) \emph{references} that are ambiguous (e.g., that could refer to multiple different entities previously mentioned in the conversation) and with a variety of reference ranges;
(ii) \emph{slot values / program arguments and functions} to be predicted that are not present verbatim in the current utterance, but require \emph{normalization} or
derivation from the dialog's context: its situation, user knowledge or world knowledge.
\bibliography{anthology,custom}

\newpage

\appendix

\section*{Appendix}
\label{sec:appendix}

\begin{table*}[ht]
\centering
\footnotesize
\begin{tabular}{l l l l}
\toprule
\textbf{}         & & \textbf{Normalized value} & \textbf{Span in utterance} \\ \midrule
\textbf{Typos} &
    & \dst{time}{18:15}         & \userspan{18:!5} \\
&    & \dst{area}{centre}        & \userspan{located in the cetre} \\
\midrule
\textbf{Entity} & 
\textbf{Alternative} 
    & \dst{area}{center}        & \userspan{centre} \\
\textbf{Recognition} & 
\textbf{Spellings} 
    & \dst{attraction}{theater} & \userspan{theatre} \\
\cmidrule{2-4}
& \textbf{Numbers}
    & \dst{price}{\$30}         & \userspan{thirty bucks} \\
&    & \dst{people}{3}           & \userspan{three} \\
\cmidrule{2-4}
& \textbf{Date \& Time}
    & \dst{time}{4:30 pm}       & \userspan{half past 4 in the evening} \\
&    & \dst{book\_stay}{7}       & \userspan{a week} \\
\cmidrule{2-4}
& \textbf{Shortcuts}
    & \dst{location}{San Francisco}        &  \userspan{San Fran}, \userspan{SF}, \userspan{SFO} \\
&    & \dst{start\_day}{saturday}           &  \userspan{sat} \\
\midrule
\textbf{Semantic} &
    & \dst{smoking\_allowed}{True}         & \userspan{smoker-friendly}, \userspan{allowed to smoke} \\
\textbf{Understanding} &
    & \dst{has\_seating\_outdoors}{True}   & \userspan{in the patio}, \userspan{Al Fresco} \\ 
&    & \dst{price\_range}{inexpensive}      & \userspan{low-cost}, \userspan{budget}, \userspan{low priced} \\
\midrule
\textbf{Computation}
&    & \dst{year}{2017}                 & \userspan{two years ago} \\
&    & \dst{book\_people}{3}            & \userspan{yes, and my 2 companions} \\
&    & \dst{book\_stay}{2}              & \userspan{from tuesday through thursday} \\
\bottomrule
\end{tabular}
\caption{\label{tbl:slot_value_normalization}Examples of value normalizations in the dialog state tracking task. The right column features spans from user or agent utterances, while the left column shows their corresponding normalized slot and value as represented in the dialog state. All examples in this table are taken from the \mwoz{} or \sgd{} datasets.}
\end{table*}


\begin{figure}[!h]
    \begin{algorithmic}
        \footnotesize
        \For{(slot, value) in turn's dialog state update}
            \While{value is not found in dialog}
                \If{value in turn(i)}
                    \State $\cdist{} \gets i$
                    \State $context \gets verbatim$
                \ElsIf{denormalized value in turn(i)}
                    \State $\cdist{} \gets i$
                    \State $context \gets normalization$
                \ElsIf{value in turn(i) with context}
                    \State $\cdist{} \gets i$
                    \State $context \gets context~type$
                \Else
                    \State $i \gets i - 1$ \Comment{rewind by 1 turn}
                \EndIf                    
            \EndWhile
        \EndFor
    \end{algorithmic}
    \caption{\label{fig:model_independent_analysis:pseudocode}
    Pseudo-code used for the model-independent dataset analysis. The procedure is applied to each dialog in a dataset's test set and each user turn of a dialog (there is no dialog state tracking for agent turns) to measure the turn's conversationality and contextuality. Slot values that can be found programmatically by generating denormalized variations and regex matching are tagged automatically, the others are inspected manually in order to identify annotation errors and contextual effects.}
\end{figure}

\begin{table*}[!ht]
\footnotesize
\centering
\begin{tabular}{@{}llS[table-format=5]S[table-format=4]S[table-format=2.2]S[table-format=2.2]S[table-format=2.2]S[table-format=2.2]S[table-format=2.2]S[table-format=2.2]S[table-format=2.2]@{}}
\toprule
\textbf{Dataset} & 
\textbf{Input} & 
\textbf{Fine-tuning} & 
\textbf{input} & 
\textbf{output} &
\textbf{training} &
\textbf{CO2 emissions} \\
 & 
\textbf{representation} & 
\textbf{steps} & 
\textbf{seq. length} & 
\textbf{seq. length} &
\textbf{time [hours]} &
\textbf{(estimate in kg)} \\
\midrule
\textbf{\mwoz{}}
& \Hquad current user turn 
& 5k  & 256   & 128  & 0.690 & 2.09 \\
& \quad $+$ last agent turn
& 5k  & 256   & 128  & 0.712 & 2.16 \\
& \quad $+$ prev. dialog state
& 5k  & 256   & 128  & 0.708 & 2.14 \\
& \Hquad full dialog history
& 5k  & 1024  & 128  & 1.216 & 3.68 \\
\midrule
\textbf{\sgd{}}
& \Hquad current user turn
& 20k & 256   & 128  & 1.093 & 3.31 \\
& \quad $+$ last agent turn
& 20k & 256   & 128  & 1.112 & 3.37 \\
& \quad $+$ prev. dialog state
& 20k & 256   & 128  & 1.149 & 3.48 \\
& \Hquad full dialog history
& 20k & 1024  & 128  & 1.328 & 4.02 \\
\midrule
\textbf{\smcalflow{}}
& \Hquad current user turn
& 5k  & 2048   & 2048  & 3.180 & 9.63 \\
& \quad $+$ last agent turn
& 5k  & 2048   & 2048  & 3.893 & 11.79 \\
& \quad $+$ prev. dialog state
& 5k  & 2048   & 2048  & 3.707 & 11.22 \\
& \Hquad full dialog history
& 5k  & 2048   & 2048  & 3.661 & 11.08 \\
\bottomrule
\end{tabular}
\caption{Hyperparameters used for training the Dialog State Tracking (DST) with T5, corresponding to results in Table~\ref{tbl:t5_experiments}. We trained on Google Cloud TPU v3 with 32 cores and followed the high estimate procedure in~\cite{Patterson2021-iu} to estimate the resulting carbon emissions.}
\label{tbl:t5_hyperparameters}
\end{table*}

\end{document}